%% file: main.tex
\documentclass[conference]{IEEEtran}
\IEEEoverridecommandlockouts

\usepackage{cite}
\usepackage{amsmath,amssymb,amsfonts}
\usepackage{graphicx} 
\usepackage{multirow} 
\usepackage{booktabs} 
\usepackage{hyperref}
\usepackage{algorithmic}
\usepackage{graphicx}
\usepackage{textcomp}
\usepackage{xcolor}
\usepackage{tikz}

\def\BibTeX{{\rm B\kern-.05em{\sc i\kern-.025em b}\kern-.08em
    T\kern-.1667em\lower.7ex\hbox{E}\kern-.125emX}}

\begin{document}

\title{Backdoor Attacks on Contrastive Continual Learning for IoT Systems}

\author{
\IEEEauthorblockN{Alfous Tim    \&   Kuniyilh Simi D}
\IEEEauthorblockA{}
}

\maketitle

\begin{abstract}
The Internet of Things (IoT) systems increasingly depend on continual learning to adapt to non-stationary environments. These environments can include factors such as sensor drift, changing user behavior, device aging, and adversarial dynamics. Contrastive continual learning (CCL) combines contrastive representation learning with incremental adaptation, enabling robust feature reuse across tasks and domains. However, the geometric nature of contrastive objectives, when paired with replay-based rehearsal and stability-preserving regularization, introduces new security vulnerabilities. 
Notably, backdoor attacks can exploit embedding alignment and replay reinforcement, enabling the implantation of persistent malicious behaviors that endure through updates and deployment cycles. This paper provides a comprehensive analysis of backdoor attacks on CCL within IoT systems. We formalize the objectives of embedding-level attacks, examine persistence mechanisms unique to IoT deployments, and develop a layered taxonomy tailored to IoT. Additionally, we compare vulnerabilities across various learning paradigms and evaluate defense strategies under IoT constraints, including limited memory, edge computing, and federated aggregation. 
Our findings indicate that while CCL is effective for enhancing adaptive IoT intelligence, it may also elevate long-lived representation-level threats if not adequately secured.
\end{abstract}


\section{Introduction}




The Internet of Things (IoT) has transformed modern computing by enabling large-scale sensing, actuation, and autonomous decision-making across domains such as smart homes, healthcare, intelligent transportation, industrial automation, and critical infrastructure \cite{atzori2010iot}. Billions of interconnected devices continuously generate high-dimensional, heterogeneous data streams. Extracting actionable intelligence from these streams increasingly relies on machine learning models deployed across device, edge, and cloud layers \cite{shi2016edge}.

Unlike traditional static machine learning systems, IoT environments are inherently non-stationary. Sensor drift, device aging, firmware upgrades, environmental changes, seasonal patterns, and evolving user behaviors induce continuous distribution shifts, commonly referred to as concept drift \cite{gama2014survey}. In industrial IoT (IIoT) settings, additional sources of non-stationarity include equipment wear, changes in production regimes, and maintenance cycles. Consequently, models trained offline quickly become outdated and require periodic adaptation.

Continual learning (CL) has become an effective approach for incremental model adaptation while preventing catastrophic forgetting. \cite{kirkpatrick2017ewc,rebuffi2017icarl,aljundi2019continual, chathoth2021federated}. By updating parameters sequentially while preserving prior knowledge, CL enables long-lived models suitable for edge and IoT deployments where full retraining may be infeasible due to bandwidth, latency, or operational constraints~\cite{chathoth2022differentially}. Replay-based methods store a subset of historical data for rehearsal, while regularization-based approaches constrain parameter updates to protect important knowledge~\cite{rolnick2019experience}.

Large language models (LLMs), along with contextual augmentation, have proven effective in improving the performance of anomaly detection~\cite{meguellati2025llm, peng2025log, ning2025user}. 
Contrastive learning in LLMs improves performance by training models to distinguish between similar (positive) and dissimilar (negative) data pairs, creating more robust, discriminative, and semantically accurate representations~\cite{khosla2020supcon}. It enhances tasks like sentence embedding, instruction tuning, and reasoning by aligning desired outputs while avoiding undesirable ones. 
The contrastive learning paradigm, together with LLMs, offers numerous possibilities for IoT security~\cite {khosla2020supcon}. Recently, contrastive learning has been integrated into continual learning frameworks to improve representation stability and transferability \cite{cha2021co2l}. Contrastive objectives, such as InfoNCE \cite{chen2020simclr} and supervised contrastive learning \cite{khosla2020supcon}, learn embedding spaces where semantically related samples cluster tightly and unrelated samples are separated. In IoT systems, where labeled data is scarce and unlabeled data is abundant, self-supervised contrastive learning is particularly attractive \cite{he2020moco, chathoth2025dynamic}. By leveraging geometric structure in representation space, contrastive continual learning (CCL) improves robustness to drift and enhances cross-task generalization.

However, the very mechanisms that make CCL effective may also introduce new security vulnerabilities~\cite{carlini2021poisoning}. IoT systems operate in distributed and often adversarial environments. Devices may be physically accessible, communication channels may be insecure, and federated learning may aggregate updates from heterogeneous clients \cite{bagdasaryan2020backdoor}. These characteristics expand the attack surface beyond traditional centralized machine learning.

Backdoor attacks are among the most insidious threats in this setting. A backdoor attack implants a hidden trigger such that a model behaves normally on clean inputs but produces attacker-controlled outputs when the trigger is present \cite{gu2017badnets, chathoth2024dynamic}. Such attacks have been demonstrated in supervised classification, federated learning, and self-supervised pretraining \cite{turner2019label,bagdasaryan2020backdoor,carlini2023poison,gao2023backdoor, chathoth2025pcap}. In IoT contexts, triggers may correspond to subtle signal perturbations, rare operational regimes, or semantic environmental conditions, making detection particularly difficult.

While prior work has analyzed backdoors in static and supervised continual learning \cite{yang2022backdoorcl}, the interaction between backdoor attacks and contrastive continual learning remains underexplored. Contrastive objectives explicitly shape the geometry of embedding space. If an adversary aligns triggered samples with target-class clusters, replay mechanisms and representation distillation may repeatedly reinforce this alignment across tasks. Unlike purely supervised settings, where backdoors are tied primarily to output-layer parameters, CCL may embed malicious associations directly into feature manifolds.

This risk is amplified in IoT deployments for several reasons. First, replay buffers are frequently used at the edge to mitigate forgetting, potentially reintroducing poisoned samples at every update cycle. Second, federated or distributed aggregation can propagate compromised representations across devices or sites. Third, IoT systems often have limited retraining windows, meaning compromised models may persist for extended periods without a full reset. Finally, resource constraints restrict the applicability of computationally heavy defense mechanisms.

The convergence of contrastive geometry, continual adaptation, and distributed IoT deployment creates a unique and under-examined security landscape. A systematic analysis of backdoor vulnerabilities in contrastive continual learning is therefore essential for ensuring the security and adaptability of IoT intelligence.

In this paper, we investigate backdoor attacks on CCL from an IoT-centric perspective. We formalize embedding-level attack objectives, analyze persistence mechanisms induced by replay and stability constraints, develop a layered taxonomy aligned with IoT architectures, and examine defense strategies under realistic edge and federated constraints. Our findings suggest that contrastive continual learning, while powerful for non-stationary IoT environments, may inadvertently amplify long-lived representation-level threats if not designed with adversarial robustness in mind.


\section{Contrastive Continual Learning in IoT}

\subsection{Machine Learning in IoT Deployment}
IoT learning architectures consist of three primary layers: the device layer, the edge layer, and the cloud layer. The device layer is responsible for data collection and lightweight inference, enabling devices to gather information and make preliminary decisions based on that data. 
Although IoT edge devices are resource-constrained, machine learning (ML)-based models can complement rule-based anomaly detection used in embedded IoT devices~\cite{melnyk2025hardware}. Such an ML-based model can leverage locally generated data to train the global model deployed in the cloud.
Building on this, the edge layer focuses on local training and utilizes replay buffers to enhance learning efficiency and responsiveness by processing data closer to the source. Finally, the cloud layer plays a crucial role in aggregating data from various devices and facilitating federated updates, thereby enabling improved learning across the entire network.
Continual updates in these architectures may occur at either the edge or cloud layer, depending on the available resources and specific operational constraints
Continual updates may occur at the edge or cloud, depending on resource constraints \cite{lourenco2025ondevice}.

\subsection{Continual Learning in IoT}

IoT systems operate in dynamic environments where data distributions evolve continuously due to sensor drift, environmental variability, device aging, software updates, and changing user behavior. Traditional static machine learning models trained offline are insufficient in such settings, as their performance degrades under distribution shifts. Continual learning (CL) provides a framework for incremental adaptation without catastrophic forgetting, enabling models to update sequentially while retaining prior knowledge.

IoT systems commonly employ continual learning under the following settings:
Domain-Incremental Learning (DIL) refers to the scenario in which the same task is performed while the sensor distributions evolve over time. In contrast, Class-Incremental Learning (CIL) involves the introduction of new event types or faults over time. Stream-Incremental Learning enables continuous updates without clear task boundaries.
Replay-based methods are especially common in IoT, where limited historical data is stored at edge devices or gateways for rehearsal. Regularization-based approaches such as EWC are also used to preserve stability under resource constraints \cite{kirkpatrick2017ewc}. Recently, contrastive and self-supervised learning have been adopted to leverage abundant unlabeled IoT data \cite{cha2021co2l}.

\subsection{Contrastive Continual Learning in IoT}

IoT systems present unique challenges for continual learning. First, labeled data is often scarce or delayed, particularly in anomaly detection or predictive maintenance applications. Second, edge devices have limited computational and memory resources, constraining model complexity and replay buffer size. Third, IoT data streams are frequently heterogeneous and multimodal, including time-series signals, images, and environmental measurements. These characteristics motivate the integration of contrastive learning into continual frameworks~\cite{li2024importance, wang2022continual_contrastive}.

Contrastive learning is particularly well-suited to IoT environments because it leverages large volumes of unlabeled data through self-supervised objectives~\cite{chathoth2026contrastive, khosla2020supcon}. By learning representations that bring semantically similar samples closer while pushing dissimilar samples apart, contrastive objectives construct structured embedding spaces without requiring extensive annotations. This property is especially valuable in IoT, where continuous data collection generates abundant raw signals but labeling is expensive or impractical.

In contrastive continual learning (CCL), the model learns sequentially over tasks or time increments while maintaining representation consistency. Let $\mathcal{D}_t$ denote data collected during increment $t$. The encoder $f_\theta$ is trained to minimize:

\begin{equation}
\mathcal{L}_{\text{CCL}} =
\mathcal{L}_{\text{contrastive}}(\mathcal{D}_t \cup \mathcal{M})
+ \lambda \mathcal{L}_{\text{distill}},
\end{equation}

where $\mathcal{M}$ is a replay memory containing samples from previous increments and $\mathcal{L}_{\text{distill}}$ preserves embedding relations learned in earlier stages.

\input{design}

\section{Defense Strategies for IoT Contrastive Continual Learning}

Defending against backdoor attacks in IoT-oriented contrastive continual learning (CCL) requires addressing vulnerabilities at multiple levels: input data, embedding space, replay memory, training dynamics, and federated aggregation. Unlike static supervised models, CCL systems continuously adapt under non-stationary data streams and resource constraints. Therefore, effective defenses must operate incrementally, efficiently, and without assuming access to large clean reference datasets.
We organize defense strategies into five categories: (1) data-level defenses, (2) embedding-level defenses, (3) replay-aware defenses, (4) training-time regularization, and (5) federated robustness mechanisms.

\subsection{Data-Level Defenses}

Data-level defenses attempt to detect and filter poisoned samples before they influence representation learning.

\subsubsection{Anomaly and Drift-Aware Filtering}

In IoT systems, poisoned samples may resemble natural distribution shifts. Therefore, standard anomaly detection must be augmented with drift-aware modeling \cite{gama2014survey}. Techniques include online statistical monitoring of feature distributions, cross-sensor consistency validation, and physics-informed constraints for industrial IoT.
However, purely statistical anomaly detection may fail against semantic or latent triggers that mimic legitimate operating conditions.

\subsubsection{Spectral and Activation-Based Detection}

Spectral signature methods \cite{tran2018spectral} analyze covariance structure in intermediate representations to identify poisoned clusters. Activation clustering \cite{chen2019activation} groups hidden-layer activations to detect abnormal class subclusters.
In CCL, these methods must operate on embedding space rather than solely on classifier outputs. Since contrastive learning shapes embedding geometry, cluster-level inspection becomes particularly important.
However, in IoT CCL, replay buffers may contain few samples, reducing statistical power for reliable cluster separation.

\subsection{Embedding-Level Defenses}

Because CCL backdoors operate primarily in embedding space, defenses must directly audit geometric structure.

\subsubsection{Cluster Geometry Auditing}

Contrastive learning produces compact class clusters. A backdoor may create an anomalous sub-cluster within a target class. Periodic cluster integrity checks can measure intra-class variance, inter-class margin stability, and outlier embedding density.
Significant deviations may indicate embedding hijacking.

\subsubsection{Trigger-Invariant Contrastive Objectives}

One proactive strategy is to modify the contrastive objective to penalize excessive concentration of similarity within small subsets of samples, as per the equation:

\begin{equation}
\mathcal{L}_{robust} =
\mathcal{L}_{contrastive}
+ \beta \, \text{ClusterRegularization}
\end{equation}

This discourages the formation of tight, malicious sub-clusters.

\subsubsection{Representation Smoothing}

Embedding smoothing techniques add stochastic perturbations to latent vectors during training. This reduces sensitivity to specific trigger alignments and weakens geometric entrenchment.

\subsection{Replay-Aware Defenses}

Replay is a core amplification mechanism in CCL, making memory sanitization essential.

\subsubsection{Memory Sanitization}

Replay samples should be periodically audited before reuse. Techniques include outlier filtering in embedding space, influence function analysis, and diversity-based memory pruning.
If a small set of samples disproportionately influences embedding alignment, they may be flagged.

\subsubsection{Adaptive Replay Sampling}

Instead of uniform sampling from memory, replay selection can incorporate uncertainty estimation, sample trust scores, and temporal-decay weighting.
In IoT systems, where memory capacity is limited, careful selection reduces replay amplification of poisoned samples.

\subsubsection{Controlled Forgetting}

Introducing mild forgetting mechanisms may prevent indefinite persistence of malicious alignments. Periodic reinitialization of projection heads or partial embedding resets can mitigate long-term entrenchment.

\subsection{Training-Time Regularization}

Training procedures can incorporate adversarial robustness mechanisms.

\subsubsection{Adversarial Unlearning}

Unlearning techniques \cite{neel2021descent} attempt to remove influence of suspected samples by reversing gradient contributions. In CCL, unlearning must operate at embedding level and may require partial retraining.

\subsubsection{Contrastive Adversarial Training}

Adversarial examples can be generated during training to test cluster robustness. If minor perturbations drastically change embedding similarity, the model may be overly sensitive to triggers.

\subsubsection{Stability–Plasticity Rebalancing}

Reducing overemphasis on stability constraints may allow natural forgetting of malicious alignments. Dynamic adjustment of distillation weights $\lambda$ can prevent over-preservation of compromised embeddings.

\subsection{Federated and Distributed Defenses}

IoT deployments frequently involve federated learning across edge devices.

\subsubsection{Robust Aggregation}

Aggregation rules robust to outliers (e.g., median-based, trimmed mean) can reduce impact of malicious clients \cite{bagdasaryan2020backdoor}.

\subsubsection{Client Reputation Systems}

Maintaining historical trust scores for clients allows dynamic weighting of contributions.

\subsubsection{Cross-Client Embedding Consistency Checks}

Since contrastive learning enforces representation alignment, cross-client embedding distribution comparisons may reveal anomalous geometric distortions introduced by malicious participants.

\subsection{IoT-Specific Deployment Constraints}

Designing defenses for IoT CCL must account for several key factors, including limited memory in replay buffers, low computational capacity at edge nodes, communication bandwidth constraints, intermittent connectivity, and scarce labeled data.
Heavyweight forensic methods may be impractical. Therefore, lightweight, incremental, and embedding-aware defenses are preferable.

\subsection{Defense Challenges Unique to CCL}

Defending CCL differs from defending static models in several respects, including continuous parameter updates, replay-induced reinforcement, representation reuse across tasks, and drift-induced ambiguity.
A key open challenge is balancing robustness with adaptability. Overly aggressive filtering may degrade continual learning performance, while weak filtering may allow persistent backdoors.
%
%
Because contrastive objectives shape geometric structure, securing IoT CCL demands moving beyond output-layer defenses toward representation-centric protection mechanisms.












\section{Open Research Directions}

Although backdoor attacks in supervised and federated learning have received increasing attention, the security of contrastive continual learning (CCL) in IoT environments remains largely unexplored. The convergence of sequential adaptation, replay-based rehearsal, embedding-level objectives, and distributed deployment introduces a uniquely persistent and complex threat model. Addressing these challenges requires advances that span theory, algorithm design, and systems engineering.

A fundamental open problem is the development of certified secure contrastive continual learning for IoT systems. Existing backdoor defenses are predominantly empirical and heuristic, offering limited guarantees on worst-case attack success rates or long-term persistence across incremental updates. In IoT deployments, where models may operate for extended periods without full retraining, formal guarantees become especially important. Future work must establish theoretical bounds on attack success rate under constrained poisoning budgets, characterize how replay influences embedding stability, and derive conditions under which malicious geometric alignment decays over time. Robust contrastive objectives with provable embedding separation margins and stability guarantees across sequential updates represent a promising direction. Without formal certification, IoT CCL systems deployed in safety-critical settings remain vulnerable to stealthy long-lived attacks.

Another pressing research direction concerns trigger-resistant replay sampling. Replay mechanisms are central to continual learning, yet they simultaneously serve as a primary amplification channel for backdoor persistence. Standard replay strategies prioritize diversity or class balance, but they rarely account for adversarial risk. In IoT environments, replay buffers are small due to memory constraints, which can inadvertently increase the influence of poisoned samples. Research is needed to design replay policies that incorporate trust scoring, influence estimation, or anomaly-aware weighting. Such mechanisms must balance two competing objectives: preserving legitimate historical knowledge while minimizing reinforcement of suspicious embeddings. Developing principled replay selection criteria that remain lightweight enough for edge deployment remains an open and challenging problem.

Adaptive forgetting under adversarial drift represents another critical area of inquiry. Continual learning traditionally seeks to minimize forgetting; however, in adversarial settings, selective forgetting may be beneficial. IoT systems experience natural concept drift caused by environmental and operational changes. Distinguishing between legitimate drift and malicious geometric shifts is nontrivial. Future research should explore adversarially-aware stability–plasticity trade-offs, where the strength of representation distillation or regularization dynamically adapts based on anomaly signals in embedding space. Controlled resetting of projection heads, partial representation reinitialization, or dynamic distillation weighting may provide mechanisms to prevent indefinite preservation of compromised embeddings. Understanding how to forget malicious structure without degrading overall performance is a core unsolved challenge.

Multi-modal IoT backdoor detection also warrants deeper investigation. Many IoT deployments integrate heterogeneous modalities such as time-series sensor data, images, audio, and environmental signals. Contrastive learning often aligns these modalities into a shared embedding space, enabling cross-modal retrieval and reasoning. However, this alignment introduces new vulnerabilities: a backdoor injected in one modality may propagate across modalities through shared representation geometry. Future research should investigate cross-modal consistency checks, disagreement detection mechanisms, and modality-aware embedding auditing. Developing robust multi-modal contrastive objectives that prevent malicious cross-modal alignment without sacrificing semantic coherence remains an open problem.

A related challenge is understanding the long-term persistence dynamics of backdoors in CCL. While empirical studies demonstrate that replay and representation distillation reinforce malicious alignment, formal models of persistence decay are lacking. It remains unclear under what conditions backdoor effects attenuate naturally as new tasks are introduced, or whether certain geometric distortions become permanently embedded in representation space. Analyzing the stability of poisoned clusters, sensitivity of embedding manifolds to distribution shifts, and interaction between replay frequency and persistence may enable principled forgetting strategies. Persistence modeling is particularly important in IoT systems where devices operate continuously and retraining opportunities are limited.

Energy- and resource-aware security mechanisms constitute another important research direction. IoT devices operate under strict constraints on memory, computational power, and communication bandwidth. Many backdoor detection techniques rely on large-scale clustering, covariance analysis, or full retraining, which may be infeasible at the edge. Lightweight embedding anomaly metrics, incremental cluster statistics, and efficient replay sanitization techniques must be developed to ensure practical deployability. Security mechanisms that impose minimal overhead while maintaining continual learning performance are essential for real-world IoT adoption.

Finally, federated continual robustness remains an open frontier. In distributed IoT systems, federated aggregation can unintentionally propagate compromised embedding structures across clients. Slow, stealthy geometric drift may evade traditional robust aggregation rules. Future work should explore embedding-level aggregation verification, client reputation systems tailored to continual updates, and geometric consistency checks across clients. Securing federated CCL requires rethinking aggregation not only in parameter space but also in representation space.

In summary, securing contrastive continual learning in IoT systems demands advances in certified robustness, replay-aware defenses, adaptive forgetting, multi-modal anomaly detection, persistence modeling, resource-efficient implementation, and federated robustness. The interplay between contrastive geometry, continual adaptation, and distributed IoT deployment creates a uniquely persistent threat landscape. Addressing these open challenges is essential for building trustworthy adaptive intelligence in next-generation IoT systems.

\section{Conclusion}


Contrastive continual learning (CCL) is a robust framework designed to enhance adaptive intelligence in Internet of Things (IoT) systems facing non-stationary environments. It combines contrastive representation learning with replay mechanisms to combat catastrophic forgetting while promoting cross-task generalization. However, this integration introduces new security vulnerabilities, particularly because CCL embeds knowledge within representation geometry rather than traditional decision boundaries. Adversaries can exploit this by aligning malicious inputs with target clusters in the embedding space, thereby establishing a persistent backdoor effect that persists beyond initial attacks and across incremental updates.

The risks associated with CCL are exacerbated in IoT contexts due to the distributed nature of device-edge-cloud architectures and replay buffers, which reinforce poisoned samples over time. This creates widespread vulnerabilities, as compromised representations can propagate through federated learning systems and remain undetected for extended periods. The study outlines a layered taxonomy of backdoor attacks and highlights the limitations of existing defense mechanisms, emphasizing the need for embedding-centric protection strategies. As CCL expands in IoT scenarios, addressing these vulnerabilities is essential for secure and trustworthy adaptive intelligence in future cyber-physical systems.

\bibliographystyle{IEEEtran}
\bibliography{bib}

\end{document}

%% file: design.tex
\section{Comparison with Standard Continual Learning}

\begin{table}[t]

\centering
\caption{Security Comparison Across Learning Paradigms}
\label{tab:MLvsCLvsCCLnew}
\begin{tabular}{lccc}
\toprule
Property & Static ML & CL & CCL \\
\midrule
Replay Amplification & No & Yes & Strong \\
Embedding Persistence & Moderate & Moderate & High \\
Latent Trigger Risk & Moderate & Moderate & High \\
Federated Vulnerability & Moderate & High & Very High \\
\bottomrule
\end{tabular}
\end{table}

A backdoor attack implants a trigger $\delta$ such that a trained model behaves normally on benign inputs but outputs an attacker-chosen prediction when the trigger is present:
\[
f_\theta(x) = y, \quad f_\theta(x \oplus \delta) = y^*
\]

In IoT, triggers may correspond to abnormal sensor patterns, specific signal perturbations, or environmental or operational conditions~\cite{gao2023backdoor, yang2022backdoorcl, bagdasaryan2020backdoor, chathoth2024dynamic, chathoth2025pcap}.
Such triggers and their objectives are designed based on the adversary's capabilities.


\textit{Adversary Capabilities:}
Table~\ref{tab:adversary_capabilities} provides the details of adversary capabilities considered in the backdoor attack on CCL.
No matching results, press enter to execute your custom prompt repalce itemize with textual details in a paragraph
In the context of security threats, several key points of compromise are noteworthy. Firstly, sensor-level compromise involves injecting poisoned measurements from compromised devices, which can lead to inaccurate data being processed. Additionally, there is the risk of edge-level access, where malicious actors may manipulate replay buffers or interfere with local model updates, ultimately affecting the integrity of the system. Furthermore, federated clients can pose a threat by participating as malicious entities in federated continual learning, undermining the collaborative learning process. Lastly, pretraining poisoning is a concern, as it entails the act of contaminating self-supervised or contrastive pretraining data, which can compromise model performance from the outset.

\textit{Adversary Objectives:}
The adversary's main objective is to preserve "Stealth", which refers to the backdoor's ability to operate without drawing attention, thereby ensuring that the system maintains normal performance during routine operations. The other factor, "Persistence," highlights the robustness of the backdoor, indicating that it can endure through system updates and adaptation processes that typically occur over time. Lastly, "Operational impact" emphasizes the potential risks associated with a backdoor, such as triggering safety violations, causing missed alarms, or leading to control failures, all of which could compromise the system's integrity and safety.

Backdoor vulnerabilities in contrastive continual learning (CCL) differ fundamentally from those in both static supervised learning and standard continual learning (CL). In this section, we analyze these differences along four key dimensions: replay amplification, representation persistence, latent trigger risk, and federated propagation. Understanding these distinctions is critical for designing secure IoT-oriented learning systems.

\subsection{Static Supervised Learning vs Continual Learning}

In static supervised learning, backdoor attacks are typically embedded during a single training phase \cite{gu2017badnets}. Once training is complete, the model is fixed. The backdoor is generally tied to specific parameters—often in the later layers—and its persistence depends primarily on the stability of the trained network.

In contrast, continual learning introduces sequential updates:
\[
\theta_{t} \rightarrow \theta_{t+1}
\]
where each increment updates parameters while attempting to preserve prior knowledge. This dynamic setting fundamentally changes backdoor behavior. While catastrophic forgetting may reduce the effectiveness of some backdoors, replay and regularization mechanisms may instead preserve malicious features \cite{yang2022backdoorcl}.

\subsection{Standard Continual Learning vs Contrastive Continual Learning}

Standard CL methods primarily operate in output space. Their objective is to preserve classification performance via parameter regularization, including Elastic Weight Consolidation (EWC), knowledge distillation, and replay of labeled samples~\cite{li2024continual, smith2023closer}.
Backdoors in such systems generally affect the decision boundaries of the output layers.
In contrast, CCL operates directly in embedding space. Contrastive objectives explicitly shape geometric relationships among samples:

\[
\mathrm{sim}(f_\theta(x_i), f_\theta(x_j))
\]

This geometric enforcement introduces deeper persistence mechanisms. If triggered samples are aligned with a target cluster, the contrastive loss continuously reinforces this alignment across updates.

\subsection{Replay Amplification}

Replay mechanisms exist in both CL and CCL. However, their effect differs.
In supervised CL, replay reinforces label associations. Poisoned samples may remain in memory, but their influence is limited to classification loss.
In CCL, replay reinforces the alignment of embeddings. Each replayed poisoned sample contributes to both positive-pair attraction and negative-pair repulsion in representation space. This geometric reinforcement strengthens cluster-level corruption, not merely output-level misclassification.
Thus, replay amplification is substantially stronger in CCL.

\subsection{Embedding Persistence / Representation Lock-In}

A central distinction lies in representation lock-in.
Standard CL preserves outputs and selected parameters, but may allow internal representations to drift if classification performance remains stable.
CCL explicitly preserves relational geometry through contrastive loss and representation distillation \cite{cha2021co2l}. This means that cluster structures are stabilized and similarity relationships are preserved, while embedding manifolds are reused across tasks.
If a backdoor alters the embedding geometry, this distortion becomes structurally embedded. Subsequent tasks inherit the compromised manifold.

\subsection{Latent Trigger Risk}

In standard CL, triggers are typically input-visible or tied to decision boundaries. Input filtering or output monitoring may detect anomalies.
In CCL, triggers may exist entirely in latent space. Because the attack objective manipulates similarity structure rather than solely classification outputs, input-level sanitization is insufficient.
Latent triggers may generalize across tasks, transfer to unseen domains, and persist even after classifier reinitialization.
This latent risk is significantly elevated in CCL.

\subsection{Federated and Distributed Vulnerability}

IoT deployments often rely on federated or distributed continual learning. In supervised CL, poisoned gradients affect classification parameters.
In CCL, poisoned gradients affect the alignment of the embedding space. Because representation learning occurs early in the network, malicious updates may propagate broadly across tasks and devices.
Federated aggregation may therefore distribute compromised embedding geometry across clients, thereby amplifying the reach of attacks.

\subsection{Impact of Non-Stationarity in IoT}

IoT environments exhibit concept drift and dynamic distributions \cite{gama2014survey}. In standard CL, drift adaptation may incidentally reduce earlier backdoor influence if representations shift.
In CCL, however, drift adaptation is constrained by geometric preservation. Contrastive objectives promote the stability of embeddings even when exposed to new data. This stability, while beneficial for transfer, may unintentionally preserve malicious alignment.
Thus, non-stationarity in IoT may mask rather than eliminate backdoor effects.

\subsection{Summary of Key Differences}

We summarize the comparison in Table~\ref{tab:MLvsCLvsCCLnew}.
In machine learning, various approaches exhibit varying degrees of persistence against backdoor attacks. Static machine learning (ML) allows for backdoors to be embedded initially, but their effectiveness relies heavily on the stability of the model's parameters over time. In standard continual learning (CL), replaying previous data helps maintain the model's output behavior, resulting in moderate persistence of the backdoor. On the other hand, CCL enhances this by employing replay and incorporating contrastive geometry, which helps to preserve the underlying structure of the embeddings. This approach leads to greater persistence of the backdoor, making it more difficult to mitigate or remove than with other methods. The geometric nature of contrastive objectives transforms backdoor attacks from surface-level output manipulation into deep representation-level entrenchment. In IoT systems—where continual adaptation, replay, and distributed aggregation are common—this creates uniquely persistent and difficult-to-detect threats.


\begin{table*}[t]
\centering
\caption{Summary of Adversary Capabilities in IoT Contrastive Continual Learning}
\label{tab:adversary_capabilities}
\renewcommand{\arraystretch}{1.15}
\begin{tabular}{p{3cm} p{3.5cm} p{3.5cm} p{4.5cm}}
\toprule
\textbf{Capability} & \textbf{Attack Vector} & \textbf{Affected Learning Component} & \textbf{IoT-Specific Impact and Persistence Mechanism} \\
\midrule

Sensor-Level Compromise &
Injection of poisoned or manipulated sensor measurements at the data source &
Contrastive embedding formation; feature extractor updates &
Malicious patterns embedded during data ingestion; difficult to distinguish from natural drift; reinforced via replay during continual updates. \\

Edge-Level Access &
Manipulation of replay buffers or local training updates at edge nodes &
Replay memory $\mathcal{M}$; incremental parameter updates $\theta_t$ &
Replay amplification strengthens poisoned embeddings across tasks; small buffer size increases influence of malicious samples. \\

Federated Client Participation &
Malicious client contributes poisoned gradients or embeddings during aggregation &
Global model aggregation; distributed representation alignment &
Compromised embedding geometry propagates across devices; stealthy drift across rounds; amplified by representation consistency enforcement. \\

Pretraining Poisoning &
Corruption of unlabeled self-supervised or contrastive pretraining data &
Initial embedding manifold initialization &
Backdoor embedded in foundational representation space; persists across all downstream continual tasks; hard to remove without full retraining. \\

\bottomrule
\end{tabular}
\end{table*}

\begin{table*}[t]
\centering
\caption{Taxonomy of Backdoor Attacks in Continual Learning for Industrial IoT Systems}
\label{tab:iiot_backdoor_taxonomy-new}
\renewcommand{\arraystretch}{1.2}
\begin{tabular}{p{3.2cm} p{3.6cm} p{4.8cm} p{4.4cm}}
\toprule
\textbf{Dimension} &
\textbf{Category} &
\textbf{Description} &
\textbf{Industrial IoT Impact} \\
\midrule

\multirow{3}{*}{\textbf{System Layer}}
& Field / Device Layer
& Poisoned sensor streams or compromised field devices inject backdoor patterns into raw measurements.
& Suppressed alarms, undetected equipment degradation, unsafe operating conditions. \\

& Edge / Gateway Layer
& Manipulation of replay buffers or incremental updates at edge gateways performing continual learning.
& Persistent backdoor reinforcement across updates; localized safety violations. \\

& Cloud / Coordination Layer
& Poisoned global model aggregation or malicious update distribution across sites.
& Cross-facility backdoor propagation; large-scale operational risk. \\

\midrule

\multirow{3}{*}{\textbf{Learning Stage}}
& Pretraining Stage
& Backdoors injected during self-supervised or contrastive representation learning.
& Deeply embedded latent triggers; difficult post-deployment removal. \\

& Incremental Update Stage
& Poisoned samples introduced during routine continual updates.
& Task- or regime-specific backdoor activation. \\

& Replay-Amplified Stage
& Backdoors reinforced via rehearsal of poisoned samples from replay buffers.
& Long-term persistence across operational phases. \\

\midrule

\multirow{3}{*}{\textbf{Trigger Type}}
& Signal-Based Triggers
& Specific time-series patterns, frequency components, or sensor correlations.
& Triggered failures under rare but realistic operating signals. \\

& Semantic / Operational Triggers
& Legitimate operating regimes (e.g., load, temperature, pressure ranges).
& Stealthy activation during critical production conditions. \\

& Latent Representation Triggers
& Manipulation of internal feature space rather than input patterns.
& Invisible triggers that evade data-level inspection. \\

\midrule

\multirow{3}{*}{\textbf{Attack Objective}}
& Targeted Misclassification
& Trigger forces incorrect prediction or fault classification.
& Missed fault detection; incorrect maintenance decisions. \\

& Representation Hijacking
& Backdoor controls feature geometry across tasks.
& Long-term degradation of diagnostic reliability. \\

& Control and Safety Manipulation
& Backdoor influences control assistance or decision-support outputs.
& Physical damage, safety incidents, regulatory violations. \\

\bottomrule
\end{tabular}
\end{table*}

\section{Taxonomy of Backdoor Attacks in IoT CCL}

Backdoor attacks in contrastive continual learning exhibit unique characteristics due to the interaction between embedding geometry, replay-based rehearsal, and distributed IoT deployment. To systematically analyze these threats, we organize them along four orthogonal dimensions: (1) system layer, (2) learning stage, (3) trigger realization, and (4) attack objective. Each dimension reflects practical vulnerabilities observed in IoT systems. We provide a summary of taxonomy in Table~\ref{tab:iiot_backdoor_taxonomy-new}.

\subsection{Dimension I: System Layer}

IoT learning pipelines typically span multiple layers—device, edge, and cloud—each introducing distinct attack surfaces.

\subsubsection{Device-Level Attacks}

At the device layer, adversaries may poison raw sensor streams. In IoT systems, data originates from physically accessible devices such as cameras, environmental sensors, smart meters, or wearable devices. Attackers may inject manipulated signals, adversarial perturbations, or synthetic patterns directly into the data acquisition process. 

In contrastive continual learning, such poisoned samples are particularly dangerous because they enter the embedding formation stage. If triggered samples are repeatedly observed, contrastive objectives may align them with specific semantic clusters. Unlike centralized datasets, IoT device-level data are often unverifiable and lack ground-truth labels, making detection challenging.

\subsubsection{Edge-Level Attacks}

Edge devices often maintain replay buffers to mitigate catastrophic forgetting. These buffers store a subset of historical samples for rehearsal during incremental updates. An attacker who gains access to edge storage or update mechanisms can insert poisoned samples into the replay memory.

Replay-amplified attacks are especially severe in CCL. Since contrastive loss enforces embedding similarity between current and replay samples, poisoned examples are reintroduced at every update cycle. This repeated reinforcement strengthens malicious associations geometrically, leading to long-term embedding and entrenchment.

\subsubsection{Cloud/Federated-Level Attacks}

In federated IoT settings, multiple edge nodes contribute updates to a central server. Malicious clients may inject poisoned gradients or manipulated embeddings into aggregation rounds. Federated backdoor attacks have been shown to be highly effective due to model averaging mechanisms.

In CCL, federated attacks may propagate malicious embedding structures across devices. Since contrastive objectives rely on representation consistency, global aggregation can distribute poisoned cluster alignments across the entire IoT network.

\subsection{Dimension II: Learning Stage}

Backdoor attacks can occur at different stages of the learning lifecycle.

\subsubsection{Pretraining-Stage Poisoning}

Many IoT systems employ self-supervised contrastive pretraining to initialize feature encoders using large volumes of unlabeled data. Poisoning during this stage embeds malicious alignment directly into the foundational representation space.
Because continual learning builds upon this initialization, early-stage backdoors may persist indefinitely. Contrastive pretraining shapes the global embedding manifold, such that subsequent supervised fine-tuning may not remove deeply embedded triggers.

\subsubsection{Incremental Update Poisoning}

In continual adaptation phases, new tasks or data segments are incorporated sequentially. Adversaries may inject poisoned samples during specific increments. 
In CCL, incremental poisoning is strengthened by replay. Once poisoned samples are stored in memory, they are repeatedly presented alongside clean data, allowing the trigger alignment to stabilize across subsequent tasks.

\subsubsection{Replay Amplification Attacks}

Replay amplification attacks explicitly target rehearsal memory. Instead of relying solely on new poisoned samples, adversaries manipulate or bias memory selection to ensure persistent retraining on triggered examples. 
In IoT deployments with limited memory, even a small fraction of poisoned samples may dominate representation alignment due to repeated exposure across updates.

\subsection{Dimension III: Trigger Realization}

Triggers in IoT CCL differ significantly from image-based patches.

\subsubsection{Signal-Based Triggers}

Signal-based triggers involve specific temporal, spectral, or multivariate sensor patterns. For example, a particular frequency component in vibration data or a rare combination of environmental measurements may act as a trigger.
Such triggers are difficult to detect because they resemble natural fluctuations or operational variations. In contrastive learning, signal-based triggers can cluster with target embeddings if aligned during training.

\subsubsection{Semantic Operational Triggers}

Semantic triggers correspond to meaningful operational contexts (e.g., high load conditions, nighttime operation, rare maintenance states). These triggers may only appear under specific environmental or temporal conditions.
Because semantic triggers align with realistic IoT states, anomaly-based detection mechanisms may fail. Moreover, continual adaptation may treat such states as legitimate shifts in distribution, further entrenching the backdoor.

\subsubsection{Latent Representation Triggers}

Latent triggers manipulate feature-space geometry rather than input-level signals. Instead of relying on visible perturbations, the adversary shapes embedding relationships such that triggered inputs lie close to target clusters.
Latent triggers are particularly dangerous in CCL because contrastive objectives explicitly preserve geometric structure across tasks. Input-level sanitization defenses may not detect such manipulations.

\subsection{Dimension IV: Attack Objectives}

The adversary’s ultimate goal may vary.

\subsubsection{Targeted Misclassification}

The most direct objective is to force a specific output label upon trigger activation. In IoT anomaly detection systems, this may correspond to suppressing alarms for critical faults.

\subsubsection{Representation Hijacking}

Rather than altering outputs directly, an attacker may distort the embedding manifold. This may degrade downstream tasks, impair transfer learning, or bias similarity search functions.
In CCL, representation hijacking is amplified because the embedding structure is continuously preserved through distillation and replay.

\subsubsection{Cross-Task Persistence}

A defining risk of CCL is cross-task persistence. Backdoors embedded in earlier tasks may survive subsequent task updates due to geometric alignment and stability constraints. 
This persistence differentiates CCL backdoors from static supervised backdoors, as malicious behavior can propagate across domains and operational phases without direct reinjection of poisoned data.

\subsection{Interplay Across Dimensions}

Importantly, these taxonomy dimensions are not independent. 
For instance, a device-level signal trigger introduced during pretraining may persist through replay amplification at the edge. Additionally, a federated client poisoning attack may produce latent representation triggers that transfer across tasks. Finally, semantic triggers aligned during incremental updates may evade anomaly-based filtering and be reinforced via replay.
The convergence of distributed IoT architecture, contrastive geometry, and continual rehearsal creates a uniquely persistent threat model. Understanding these interactions is critical for designing secure CCL frameworks.
Table~\ref{tab:iot_taxonomy} provides a taxonomy of backdoor attacks on different stages in IoT CCL, with each backdoor's objectives.

\begin{table*}[t]
\centering
\caption{IoT-Oriented Taxonomy of Backdoor Attacks in CCL}
\label{tab:iot_taxonomy}
\begin{tabular}{p{2.5cm} p{3cm} p{4cm} p{4cm}}
\toprule
Stage & Category & IoT Mechanism & Objective \\
\midrule

System Layer &
Device & Sensor poisoning & Hard-to-detect data drift \\
&
Edge & Replay buffer manipulation & Repeated retraining \\
&
Cloud & Federated poisoning & Cross-site propagation \\

\midrule
Attack Stage &
Pretraining & Poison unlabeled data & Embedding initialization bias \\
&
Incremental Update & Task-specific poisoning & Replay reinforcement \\
&
Replay Amplification & Memory corruption & Long-term persistence \\

\midrule
Trigger Type &
Signal-based & Temporal/spectral patterns & Natural-looking anomalies \\
&
Semantic & Operational condition triggers & Rare but realistic contexts \\
&
Latent & Embedding-level manipulation & Evades input filtering \\

\bottomrule
\end{tabular}
\end{table*}